\documentclass[10pt,twocolumn,letterpaper]{article}

\usepackage{color}
\definecolor{green}{rgb}{0, 0.5, 0}
\definecolor{orange}{rgb}{0.8, 0.6, 0.2}
\definecolor{red}{rgb}{1.0, 0.0, 0.0}
\definecolor{teal}{rgb}{0.0, 0.4, 0.4}
\definecolor{purple}{rgb}{0.65,0,0.65}
\definecolor{saffron}{rgb}{0.95,0.75,0.2}
\definecolor{turquoise}{rgb}{0.0,0.5,0.5}
\definecolor {mygray}{gray}{.9}

\usepackage{booktabs}
\usepackage{colortbl}
\usepackage{makecell}
\usepackage{multirow}
\usepackage{array}
\usepackage{gensymb}
\newcolumntype{L}[1]{>{\raggedright\arraybackslash}p{#1}}
\newcolumntype{C}[1]{>{\centering\arraybackslash}p{#1}}
\newcolumntype{R}[1]{>{\raggedleft\arraybackslash}p{#1}}

\usepackage{amsmath}

\usepackage{iccv}
\usepackage{times}
\usepackage{epsfig}
\usepackage{graphicx}
\usepackage{amsmath}
\usepackage{amssymb}

\usepackage[pagebackref=true,breaklinks=true,letterpaper=true,colorlinks,bookmarks=false]{hyperref}

\iccvfinalcopy 

\def\iccvPaperID{3735} 

\ificcvfinal\fi
\begin{document}

\title{DeepICP: An End-to-End Deep Neural Network for 3D Point Cloud Registration}

\author{Weixin Lu \hspace{0.5cm} Guowei Wan \hspace{0.5cm} Yao Zhou \hspace{0.5cm} Xiangyu Fu \hspace{0.5cm} Pengfei Yuan \hspace{0.5cm} Shiyu Song\thanks{Author to whom correspondence should be addressed}\\
	Baidu Autonomous Driving Technology Department (ADT)\\
	{\tt\small \{luweixin, wanguowei, zhouyao, fuxiangyu, yuanpengfei, songshiyu\}@baidu.com}
}

\maketitle

\begin{abstract}
We present DeepICP - a novel end-to-end learning-based 3D point cloud registration framework that achieves comparable registration accuracy to prior state-of-the-art geometric methods. Different from other keypoint based methods where a RANSAC procedure is usually needed, we implement the use of various deep neural network structures to establish an end-to-end trainable network. Our keypoint detector is trained through this end-to-end structure and enables the system to avoid the inference of dynamic objects, leverages the help of sufficiently salient features on stationary objects, and as a result, achieves high robustness. Rather than searching the corresponding points among existing points, the key contribution is that we innovatively generate them based on learned matching probabilities among a group of candidates, which can boost the registration accuracy. Our loss function incorporates both the local similarity and the global geometric constraints to ensure all above network designs can converge towards the right direction. We comprehensively validate the effectiveness of our approach using both the KITTI dataset and the Apollo-SouthBay dataset. Results demonstrate that our method achieves comparable or better performance than the state-of-the-art geometry-based methods. Detailed ablation and visualization analysis are included to further illustrate the behavior and insights of our network. The low registration error and high robustness of our method makes it attractive for substantial applications relying on the point cloud registration task.
\end{abstract}
\section{Introduction}
\label{section:intro}

Point cloud registration is a task that aligns two or more different point clouds collected by LiDAR (Light Detection and Ranging) scanners by estimating the relative transformation between them. It is a well-known problem and plays an essential role in many applications, such as LiDAR SLAM \cite{zhang2014loam, deschaud2018imls, ji2018cpfg, neuhaus2018mc2slam}, 3D reconstruction and mapping \cite{shiratori2015efficient, droeschel2018efficient, yang2018robust, ding2019deepmapping}, positioning and localization \cite{Mita2014, Hamada2015, wan2018robust, Lu2019L3Net}, object pose estimation \cite{wong2017segicp} and so on.

This problem is challenging due to several aspects that are considered unique for LiDAR point clouds, including the local sparsity, the large amount of data and the noise caused by dynamic objects. Compared to the image matching problem, the sparsity of the point cloud makes finding two exact matching points from the source and target point clouds usually infeasible. It also increases the difficulty of feature extraction due to the large appearance difference of the same object viewed by a laser scanner from different perspectives. The millions of points produced every second requires highly efficient algorithms and powerful computational units. Appropriate handling of the interference caused by the noisy points of dynamic objects typically is crucial for delivering an ideal estimation.

\begin{figure}[!!t]
	\centering
	\includegraphics[width=\linewidth]{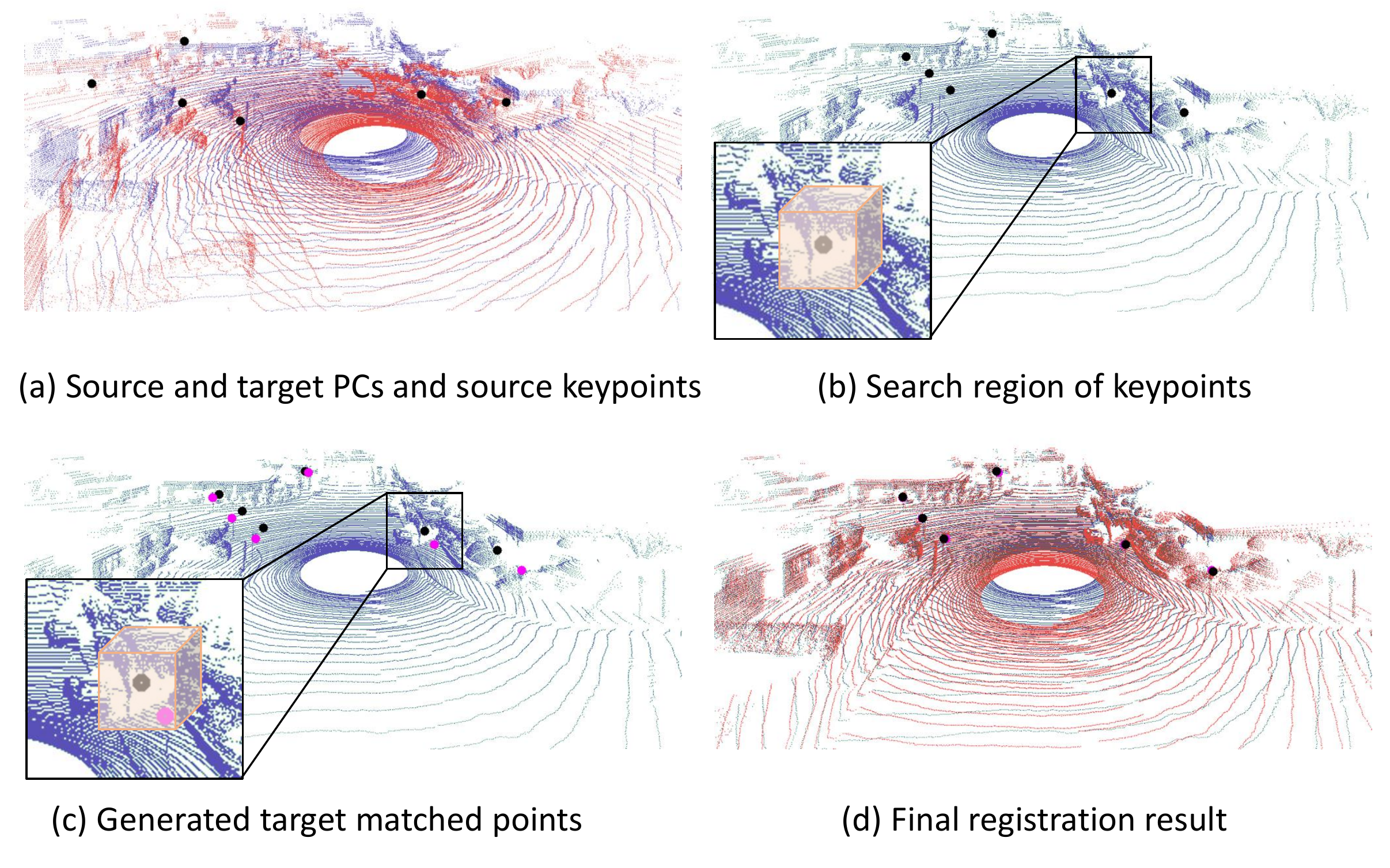}
	\caption{
		The illustration of the major steps of our proposed end-to-end point cloud registration method:
		(a) The source (red) and target (blue) point clouds and the keypoints (black) detected by the point weighting layer.
		(b) A search region is generated for each keypoint and represented by grid voxels.
		(c) The matched points (magenta) generated by the corresponding point generation layer.
		(d) The final registration result computed by performing SVD given the matched keypoint pairs.
	}
	\label{fig:intro}
	\vspace{-0.4cm}
\end{figure}

Moreover, the unbounded variety of the scene is considered as the most significant challenge in solving this problem. Traditionally, a classic registration pipeline usually includes several steps with certain variations, for example, keypoint detection, feature descriptor extraction, feature matching, outlier rejection, and transformation estimation. Although good in performance; accuracy and robustness have been achieved in some scenarios after decades of considerable engineering efforts.
Even so, finding a universal point cloud registration solution is still considered a popular, unresolved problem by the community.


Advances in deep learning have led to compelling improvements for most semantic computer vision tasks, such as classification, detection or segmentation. People are surprised by the remarkable capability of the DNNs in how well it can generalize in solving these empirically defined problems. For another important category of the problem, the geometric problems that are defined theoretically, there has been exciting progress for problems, such as stereo matching \cite{yin2018hierarchical, cheng2018learning}, depth estimation and SFM \cite{ummenhofer2017demon, zhou2018deeptam}.  Unfortunately, for 3D data, the experiential solutions of most of these attempts \cite{zeng20163dmatch, elbaz20173d, ppfnet_2018} have not been adequate enough in terms of local registration accuracy, partially due to the characteristics of the 3D point cloud mentioned above.


In this work, we propose an end-to-end learning-based method to accurately align two different point clouds. An overview of our framework is shown in Figure~\ref{fig:intro}. We name it ``DeepICP'' because Iterative Closest Point (ICP) \cite{Besl1991} is a classic algorithm that sometimes can represent the point cloud registration problem itself, and our approach is quite similar to it during the network training stage, despite the fact that there is only one iteration in the inference stage.

We first extract semantic features of each point both from the source and target point clouds using the latest point cloud feature extraction network, PointNet++ \cite{qi_2017_pointnetplusplus}.
They are expected to have certain semantic meanings to empower our network to avoid dynamic objects and focus on those stable and unique features that are good for registration.
To further achieve this goal, we select the keypoints in the source point cloud that are most significant for the registration task by making use of a point weighting layer to assign matching weights to the extracted features through a learning procedure.
To tackle the problem of local sparsity of the point cloud, we propose a novel corresponding point generation method based on a feature descriptor extraction procedure using a mini-PointNet \cite{qi2017pointnet} structure. We believe that it is the key contribution to enhance registration accuracy.
Finally, besides only using the L1 Euclidean distance between the source keypoint and the generated corresponding point as the loss, we propose to construct another corresponding point by incorporating the keypoint weights adaptively and executing a single optimization iteration using the newly introduced SVD operator in TensorFlow. The L1 Euclidean distance between the keypoint and this newly generated corresponding point is again used as another loss.
Unlike the first loss using only local similarity, this newly introduced loss builds the unified geometric constraints among local keypoints. The end-to-end closed-loop training allows the DNNs to generalize well and select the best keypoints for registration.

To summarize, our main contributions are:
\vspace{-0.2cm}
\begin{itemize}
	\item To the best of our knowledge, the first end-to-end learning-based point cloud registration framework yielding comparable results to prior state-of-the-art geometric ones.
	\vspace{-0.3cm}
	\item Our learning-based keypoint detection, novel corresponding point generation method and the loss function that incorporates both the local similarity and the global geometric constraints to achieve high accuracy in the learning-based registration task.
	\vspace{-0.3cm}
	\item Rigorous tests and detailed ablation analysis using the KITTI \cite{geiger2012we} and Apollo-SouthBay \cite{Lu2019L3Net} datasets to fully demonstrate the effectiveness of the proposed method.
\end{itemize}

\section{Related Work}
\label{section:related}


The survey work from F. Pomerleau et al. \cite{pomerleau2015review} provides a good overview of the development of traditional point cloud registration algorithms. A discussion of the full literature of the these methods is beyond the scope of this work.


The attempt of using learning based methods starts by replacing each individual component in the classic point cloud registration pipeline.
S. Salti et al. \cite{salti2015learning} proposes to formulate the problem of 3D keypoint detection as a binary classification problem using a pre-defined descriptor, and attempts to learn a Random Forest  \cite{breiman2001random} classifier that can find the appropriate keypoints that are good for matching.
M. Khoury et al. \cite{khoury2017learning} proposes to first parameterize the input unstructured point clouds into spherical histograms, then a deep network is trained to map these high-dimensional spherical histograms to low-dimensional descriptors in Euclidean space.
In terms of the method of keypoint detection and descriptor learning, the closest work to our proposal is \cite{yew20183dfeat}. Instead of constructing an End-to-End registration framework, it focuses on joint learning of keypoints and descriptors that can maximize local distinctiveness and similarity between point cloud pairs.
G. Georgakis et al. \cite{georgakis2018end} solves a similar problem for RGB-D data. Depth images are processed by a modified Faster R-CNN architecture for joint keypiont detection and descriptor estimation.
Despite the different approaches, they all focus on the representation of the local distinctiveness and similarity of the keypoints. During keypoint selection, content awareness in real scenes is ignored due to the absence of the global geometric constraints introduced in our end-to-end framework. As a result, keypoints on dynamic objects in the scene cannot be rejected in these approaches.

Some recent works \cite{zeng20163dmatch, elbaz20173d, ppfnet_2018, angelina2018pointnetvlad} propose to learn 3D descriptors leveraging the DNNs, and attempt to solve the 3D scene recognition and re-localization problem, in which obtaining accurate local matching results is not the goal. In order to achieve that, methods, as ICP, are still necessary for the registration refinement.

M. Velas et al. \cite{velas2018cnn} encodes the 3D LiDAR data into a specific 2D representation designed for multi-beam mechanical LiDARs. CNNs is used to infer the 6 DOF poses as a classification or regression problem. An IMU assisted LiDAR odometry system is built upon it. Our approach processes the original unordered point cloud directly and is designed as a general point cloud registration solution.

\begin{figure*}[!htbp]
	\centering
	\includegraphics[width=\linewidth]{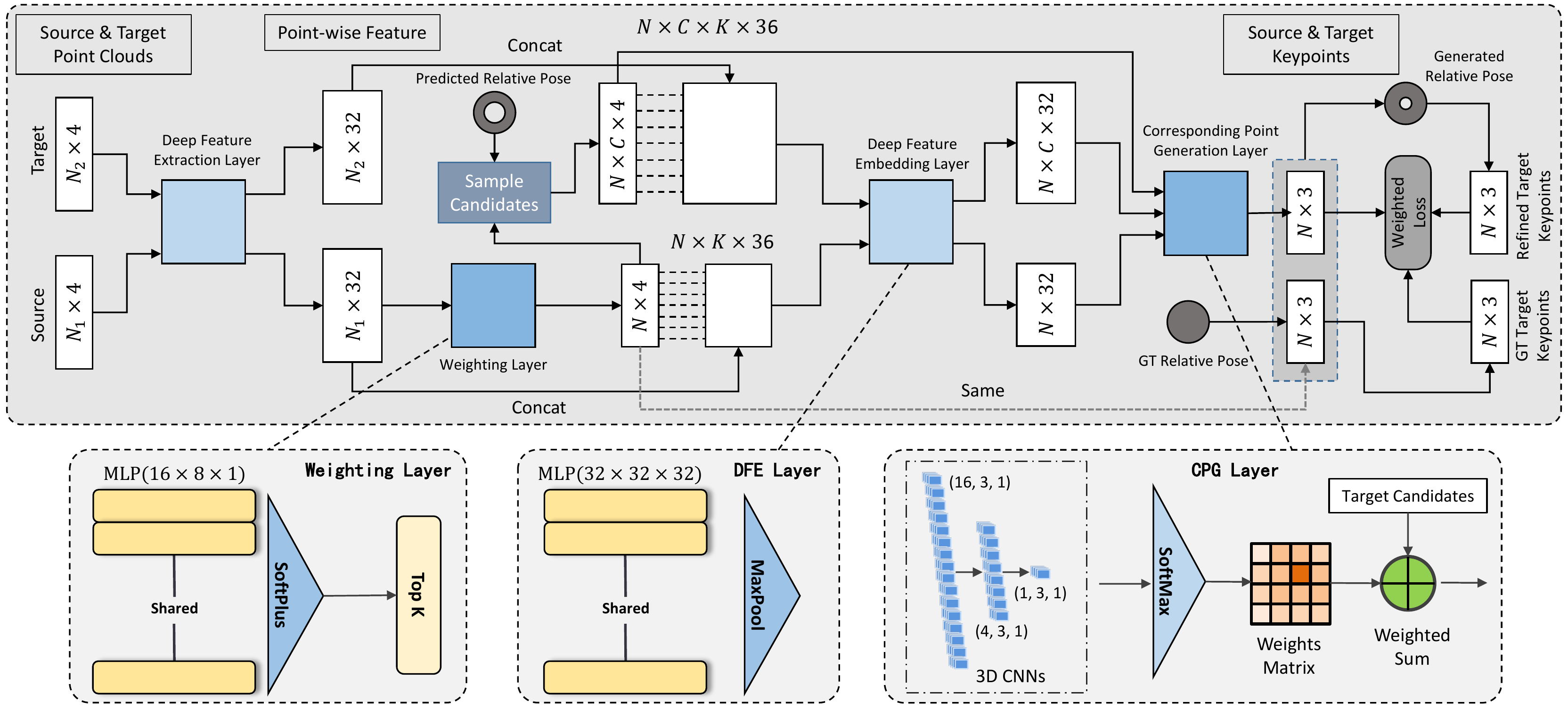}
	\caption{The architecture of the proposed end-to-end learning network for 3D point cloud registration, DeepICP. The source and target point clouds are fed into the deep feature extraction layer, then $N$ keypoints are extracted from the source point cloud by the weighting layer. $N \times C$ candidate corresponding points are selected from the target point cloud, followed by a deep feature embedding operation. The corresponding keypoints in the target point cloud are generated by the corresponding points generation layer. Finally, we propose to use the combination of two losses those encode both the global geometric constraints and local similarities.}
	\label{fig:network}
	\vspace{-0.4cm}
\end{figure*}
\section{Method}
\label{section:approach}

This section describes the architecture of the proposed network designed in detail as shown in Figure~\ref{fig:network}.

\subsection{Deep Feature Extraction}
\label{subsec:fl}

The input of our network consists of the source and target point cloud, the predicted (prior) transformation, and the ground truth pose required only during the training stage.
The first step is extracting feature descriptors from the point cloud. In the proposed method, we extract feature descriptors by applying a deep neural network layer, denoted as the Feature Extraction (FE) Layer.
As shown in Figure~\ref{fig:network}, we feed the source point cloud, represented as an $N_1 \times 4$ tensor, into the FE layer. The output is an $N_1 \times 32$ tensor representing the extracted local feature.
The FE layer here we used is PointNet++ \cite{qi_2017_pointnetplusplus} (see details in Section~\ref{section:impl}), which is a poineer work addressing the issue of consuming unordered points in a network architecture.

These local features are expected to have certain semantic meanings. Working together with the weighting layer to be introduced next, we expect our end-to-end network to be capable to avoid the interference from dynamic objects and deliver precise registration estimation. In Section~\ref{subsec:visulaizations}, we visualize the selected keypoints and demonstrate that the dynamic objects are successfully avoided.

\subsection{Point Weighting}
\label{subsec:weighting}

Inspired by the attention layer in 3DFeatNet \cite{yew20183dfeat}, we design a point weighting layer to learn the saliency of each point in an end-to-end framework. Ideally, points with invariant and distinct features on static objects should be assigned higher weights.

As shown in Figure~\ref{fig:network}, $N_1 \times 32$ local features from the source point cloud are fed into the point weighting layer. 
The weighting layer consists of a multi-layer perceptron (MLP) of 3 stacking fully connected layers and a top k operation.
The first two fully connected layers use the batch normalization and the ReLU activation function, and the last layer omits the normalization and applies the \textit{softplus} activation function.
The most significant $N$ points are selected as the keypoints through the top k operator and their learned weights are used in the subsequent processes.

Our approach is different from 3DFeatNet \cite{yew20183dfeat} in a few ways. First, the features used in the attention layer are extracted from local patches, while ours are semantic features extracted directly from the point cloud.
We have greater receptive fields learned from an encoder-decoder style network (PointNet++ \cite{qi_2017_pointnetplusplus}).
Moreover, our weighting layer does not output a 1D rotation angle to determine the feature direction, because our design of the feature embedding layer in the next section uses a symmetric and isotropic network architecture.

\subsection{Deep Feature Embedding}
\label{subsec:dfe}

After extracting $N$ keypoints from the source point cloud, we seek to find the corresponding points in the target point cloud for the final registration.
In order to achieve this, we need a more detailed feature descriptor that can better represent their geometric characteristics.
Therefore, we apply a deep feature embedding (DFE) layer on their neighborhood points to extract these local features.
The DFE layer we used is a mini-PointNet \cite{qi2017pointnet, ppfnet_2018, Lu2019L3Net} structure.

Specifically, we collect $K$ neighboring points within a certain radius $d$ of each keypoint. In case that there are less than $K$ neighboring points, we simply duplicate them.
For all the neighboring points, we use their local coordinates and normalize them by the searching radius $d$.
Then, we concatenate the FE feature extracted in Section~\ref{subsec:fl} with the local coordinates and the LiDAR reflectance intensities of the neighboring points as the input to the DFE layer.

The mini-PointNet consists of a multi-layer perceptron (MLP) of 3 stacking fully connected layers and a \textit{max-pooling} layer to aggregate and obtain the feature descriptor.
As shown in Figure~\ref{fig:network}, the input of the DFE layer is an $N \times K \times 36$ vector, which refers to the local coordinate, the intensity, and the $32$-dimensional FE feature descriptor of each point in the neighborhood.
The output of the DFE layer is again a $32$-dimensional vector.
In Section~\ref{subsec:ablation}, we show the effectiveness of the DFE layer and how it help improve the registration precision significantly.

\subsection{Corresponding Point Generation}
\label{subsec:corrs}

Similar to ICP, our approach also seeks to find corresponding points in the target point cloud and estimate the transformation. The ICP algorithm chooses the closest point as the corresponding point. This prohibits backpropagation as it is not differentiable. Furthermore, there are actually no exact corresponding points in the target point cloud to the source due to its sparsity nature. To tackle the above problems, we propose a novel network structure, the corresponding point generation (CPG) layer, to generate corresponding points from the extracted features and the similarity represented by them.

We first transform the keypoints from the source point cloud using the input predicted transformation.
Let $\{x_i, x'_i\}, i=1, \cdots, N$ denote the 3D coordinate of the keypoint from the source point cloud and its transformation in the target point cloud, respectively.
In the neighborhood of $x'_i$, we divide its neighboring space into $(\frac{2r}{s}+1, \frac{2r}{s}+1, \frac{2r}{s}+1)$ 3D grid voxels, where $r$ is the searching radius and $s$ is the voxel size.
Let us denote the centers of the 3D voxels as $\{y'_j\}, j = 1, \cdots, C$, which are considered as the candidate corresponding points.
We also extract their DFE feature descriptors as we did in Section \ref{subsec:dfe}. The output is an $N \times C \times 32$ tensor.
Similar to \cite{Lu2019L3Net}, those tensors representing the extracted DFE features descriptors from the source and target are fed into a three-layer 3D CNNs, followed by a \textit{softmax} operation, as shown in Figure~\ref{fig:network}.
The 3D CNNs can learn a similarity distance metric between the source and target features, and more importantly, it can smooth (regularize) the matching volume and suppress the matching noise.
The \textit{softmax} operation is applied to convert the matching costs into probabilities.

Finally, the target corresponding point $y_i$ is calculated through a \textit{weighted-sum} operation as:

\begin{equation} \label{equ:corresponding}
y_i=\frac{1}{\sum_{j=1}^{C}w_j}\sum_{j=1}^{C}w_j \cdot y'_j,
\end{equation}

where $w_j$ is the similarity probability of each candidate corresponding point $y'_j$. The computed target corresponding points are represented by a $N \times 3$ tensor.

Compared to the traditional ICP algorithm that relied on the iterative optimization or the methods \cite{fpfh2009icra, ppfnet_2018, zeng20163dmatch} which search the corresponding points among existing points from the target point cloud and use RANSAC to reject outliers, our approach utilizes the powerful generalization capability of CNNs in similarity learning, to directly ``guess'' where the corresponding points are in the target point cloud. This eliminates the use of RANSAC, reduces the iteration times to 1, significantly reduces the running time, and achieves fine registration with high precision.

\subsection{Loss}
\label{subsec:loss}

For each keypoint $x_i$ from the source point cloud, we can calculate its corresponding ground truth $\bar{y}_i$ with the given ground truth transformation $(\bar{R}, \bar{T})$.
Using the estimated target corresponding point $y_i$ in Section~\ref{subsec:corrs}, we can directly compute the $L1$ distance in the Euclidean space as a loss:

\begin{equation} \label{equ:loss1}
Loss_1=\frac{1}{N}\sum_{i=1}^{N}|\bar{y}_i - y_i|.
\end{equation}

If only the $Loss_1$ in Equation~\ref{equ:loss1} is used, the keypoint matching procedure during the registration is independent for each one.
Consequently, only the local neighboring context is considered during the matching procedure, while the registration task is obviously constrained with a global geometric transform.
Therefore, it's essential to introduce another loss including global geometric constraints.

Inspired by the iterative optimization in the ICP algorithm, we perform a single optimization iteration. That is, we perform a singular value decomposition (SVD) step to estimate the relative transformation $(R, T)$ given the corresponding keypoint pairs $\{x_i, y_i\}$, $i=1, \cdots, N$. Then the second loss in our network is defined as:

\begin{equation} \label{equ:loss2}
Loss_2=\frac{1}{N}\sum_{i=1}^{N}|\bar{y}_i - (Rx_i + T)|.
\end{equation}

Thanks to \cite{IonescuVS15}, the latest Tensorflow has supported the SVD operator and its backpropagation. This ensures that the proposed network can be trained in an end-to-end pattern.
As a result, the combined loss is defined as:

\begin{equation} \label{equ:loss}
Loss=\alpha Loss_1 + (1-\alpha) Loss_2,
\end{equation}
where $\alpha$ is the balancing factor. 
In Section~\ref{subsec:ablation}, we demonstrate the effectiveness of our loss design.

It's worth to note that the estimated corresponding keypoints $y_i$ are actually constantly being updated together as the estimated transformation $(R, T)$ during the training. When the network converges, the estimated corresponding keypoints become unlimitedly close to the ground truth. It's interesting that this training procedure is actually quite similar to the classic ICP algorithm. While the network only needs a single iteration to find the optimal corresponding keypoint and then estimate the transformation during inference, which is very valuable.

\section{Implementation Details}
\label{section:impl}

In the FE layer, a simplified PointNet++ is applied, in which only three \textit{set abstraction} layers with a single \textit{scale grouping} layer are used to sub-sample points into groups with sizes 4096, 1024, 256, and the MLPs of three hierarchical PointNet layer are $32 \times 32$, $32 \times 64$, $64 \times 64$ in the sub-sampling stage, and $64 \times 64$, $32 \times 32$, $32 \times 32 \times 32$ in the up-sampling stage.
This is followed by a fully connected layer with 32 kernels and a dropout layer with the keeping probability as $0.7$ to avoid overfitting.
The MLP in the point weighting layer is $16 \times 8\times 1$, and only the top $N = 64$ points are selected in the source point cloud according to their learned weights in the descending order.
The searching range $d$ and the number of neighboring points $K$ to be collected in the DFE step are set to be $1m$ and $32$, respectively.
In the mini-PointNet structure of the DFE layer, the MLP is $32 \times 32 \times 32$.
The 3D CNNs settings in the CPG step are Conv3d (16, 3, 1) - Conv3d(4, 3, 1) - Conv3d (1, 3, 1).
The grid voxels are set as $(\frac{2\times2.0}{0.4} + 1, \frac{2\times2.0}{0.4} + 1, \frac{2\times2.0}{0.25} + 1)$.

The proposed network is trained with the batch size as $1$, the learning rate as $0.01$ and the decay rate as $0.7$ with the decay step to be $10000$.
During the training stage, we conduct the data augmentation and supervised training by adding a uniformly distributed random noise of $[0.0 \sim 1.0] m$ in the $x$, $y$ and $z$ dimensions, and $[0 \sim 1.0]^\circ$ in the $roll$, $yaw$, and $pitch$ dimensions to the given ground truth.
We randomly divide the dataset into the training and validation set, yielding the ratio of training to validation as 4 to 1.
We stop at 200 epochs when there is no performance gain.

Another implementation detail worth mentioning is that we conduct a bidirectional matching strategy during inference to improve the registration accuracy. That is, the input point cloud pair is considered as the source and target simultaneously. While we don't do this during training, because this does not improve the overall performance of the model.

Moreover, all the settings above are designated for the datasets (both the KITTI and the Apollo-SouthBay) collected with Velodyne HDL64 LiDAR. 
Because the point clouds from Velodyne HDL64 are distributed within a relatively narrow region in the $z$-direction, the keypoints constraining the $z$-direction are usually quite different from the other two, such as the points on the ground plane.
This causes the registration precision at the $z$, $roll$ and $pitch$ directions to decline.
To tackle this problem, we actually duplicate the whole network structure as shown in Figure~\ref{fig:network}, and use two copies of the network in a cascade pattern.
The back network uses the estimated transformation from the front network as the input, but replaces the 3D CNNs in the CPG step of the latter with a 1D one sampling in the $z$ direction only.
Both the networks share the same FE layer, becasue we do not want to extract FE features twice.
This increases the $z$, $roll$ and $pitch$'s estimation precision.

\section{Experiments}
\label{section:exp}


\subsection{Benchmark Datasets}
We evaluate the performance of the proposed network using 11 training sequences of the KITTI odometry dataset \cite{geiger2012we}. The KITTI dataset contains point clouds captured with a Velodyne HDL64 LiDAR in Karlsruhe, Germany together with the ``ground truth'' poses provided by a high-end GNSS/INS integrated navigation system. We split the dataset into two groups, the training, and the testing. The training group includes 00-07 sequences, and the testing includes 08 - 10 sequences.

Another dataset that is used for evaluation is the Apollo-SouthBay dataset \cite{Lu2019L3Net}. It collected point clouds using the same model of LiDAR as the KITTI dataset, but, in the San Francisco Bay area, United States. Similar to KITTI, it covers various scenarios including residential areas, urban downtown areas, and highways. We also find that the ``ground truth'' poses in Apollo-SouthBay is more accurate than KITTI odometry dataset. Some ground truth poses in KITTI involve larger errors, for example, the first 500 frames in Sequence 08. Moreover, the mounting height of the LiDAR in Apollo-SouthBay is slightly higher than KITTI. This allows the LiDAR to see larger areas in the $z$ direction. We find that the keypoints picked up in these high regions sometimes are very helpful for registration. The setup of the training and test sets is similar to \cite{Lu2019L3Net} with the mapping portion discarded. There is no overlap between the training and testing data.
Refer to the supplemental material for additional experimental results using more challenging datasets.

The initial poses are generated by adding random noises to the ground truth.
In KITTI and Apollo-SouthBay, we added a uniformly distributed random error of [0 $\sim$ 1.0]$m$ in $x$-$y$-$z$ dimension, and a random error of [0 $\sim$ 1.0]$^\circ$ in $roll$-$pitch$-$yaw$ dimension.
The models in different datasets are trained separately.
Refer to the supplemental material where we evaluate robustness given inaccurate initial poses using other datasets.

\subsection{Performance}
\label{subsec:performance}


\textbf{Baseline Algorithms}
We present extensive performance evaluation by comparing with a few point cloud registration algorithms based on geometry.
They are: (i) The ICP family, such as ICP \cite{Besl1991}, G-ICP \cite{Segal2009}, and AA-ICP \cite{Pavlov2018}; (ii) NDT-P2D \cite{Todor2012}; (iii) GMM family, such as CPD \cite{Myronenko2010}; (iv) The learning-based method, 3DFeat-Net \cite{yew20183dfeat}.
The implementations of ICP, G-ICP, AA-ICP, and NDT-P2D are from the Point Cloud Library (PCL) \cite{Rusu_ICRA2011_PCL}.
Gadomski`s implementation \cite{gadomski_cpd} of the CPD method is used and the original 3DFeat-Net implementation with RANSAC for the registration task is used.

\vspace{0.2cm}
\textbf{Evaluation Criteria}
The evaluation is performed by calculating the angular and translational error of the estimated relative transformation $(R, T)$ against the ground truth $(\bar{R}, \bar{T})$.
The \textit{chordal} distance \cite{Hartley2013} between $R$ and $\bar{R}$ is calculated via the Frobenius norm of the rotation matrix, denoted as $||R - \bar{R}||_F$.
The angular error $\theta$ then can be calculated as $\theta = 2\sin^{-1}(\frac{||R - \bar{R}||_F}{\sqrt{8}})$.
The translational error is calculated as the Euclidean distance between $T$ and $\bar{T}$.

\begin{figure}[htbp]
	\centering
	\includegraphics[width=\linewidth]{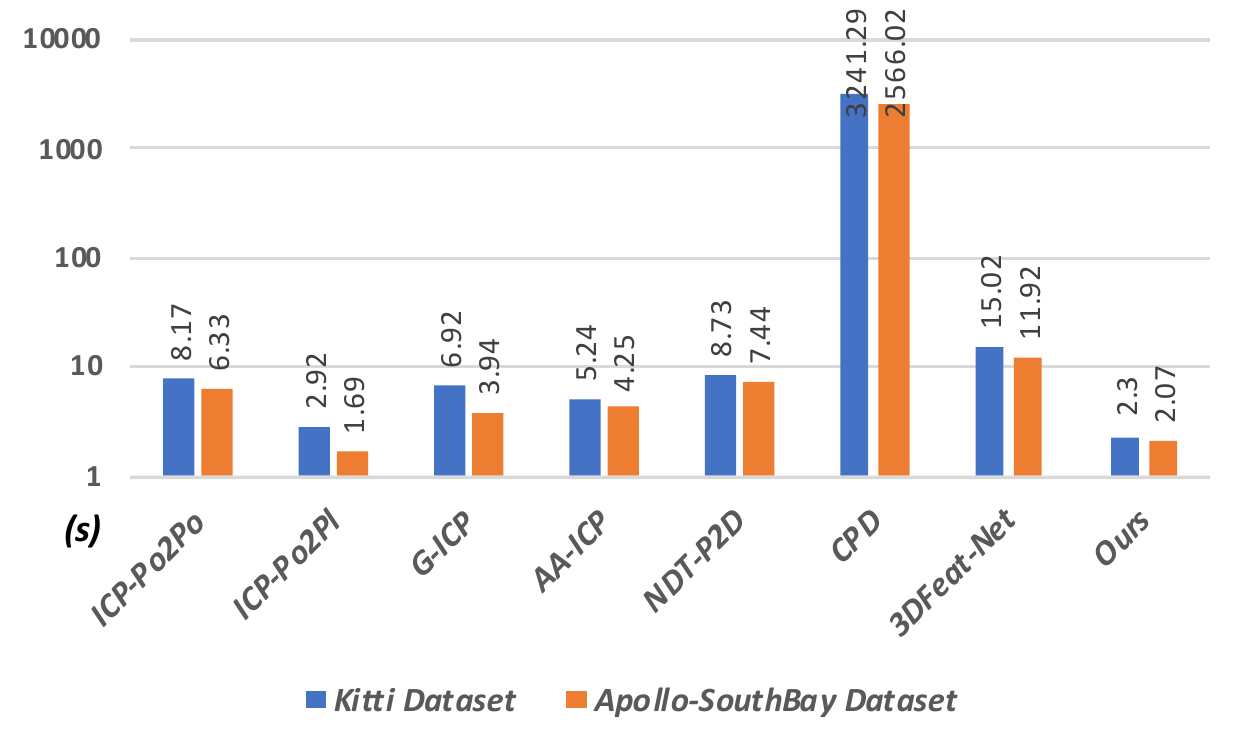}
	\vspace{-0.4cm}
	\caption{
		The running time performance analysis of all the methods. The total end-to-end inference time of our network is about $2$ seconds for registering a frame pair.
	}
	\label{fig:runtime}
	\vspace{-0.4cm}
\end{figure}

\vspace{0.2cm}
\textbf{KITTI Dataset}
We sample the input source LiDAR scans at $30$ frame intervals and enumerate its registration target within $5m$ distance to it.
The original point cloud in the dataset includes about $108,000$ points/frame.
We use original point clouds for methods such as ICP, G-ICP, AA-ICP, NDT, and 3DFeat-Net.
To keep CPD`s computing time not intractable, we downsample the point clouds using a voxel size of $0.1m$ leaving about $50,000$ points on average.
The statistics of the running time of all the methods are shown in Figure~\ref{fig:runtime}.
For our proposed method, we evaluate two versions. One is the base version, denoted as ``Ours-Base'', that infers all the degree of freedoms $x$, $y$, $z$, $roll$, $pitch$, and $yaw$ at once.
The other is an improved version with network duplication as we discussed in Section~\ref{section:impl}, denoted as ``Ours-Duplication''.
The angular and translational errors of all the methods are listed in Table~\ref{tab:kitti_data_result}.
As can be seen, for the KITTI dataset, DeepICP achieves comparable registration accuracy with regards to most geometry-based methods like AA-ICP, NDT-P2D, but performs slightly worse than G-ICP and ICP, especially for the angular error.
The lower maximum angular and translational errors show that our method has good robustness and stability, therefore it has good potential in significantly improving the overall system performance for large point cloud registration tasks.

\vspace{-0.2cm}
\begin{table}[htbp]
	\small
	\begin{center}
		\begin{tabular}{p{2.3cm}|C{1.0cm}C{1.0cm}|C{1.0cm}C{1.0cm}}
			\toprule[1pt]
			\multirow{2}{*}{\makecell[tl]{Method}} &\multicolumn{2}{c}{Angular Error($^\circ$)} &\multicolumn{2}{c}{Translation Error($m$)}  \\
			\cline{2-3}\cline{4-5}
			& Mean & Max & Mean & Max  \\
			\midrule[.5pt]
			ICP-Po2Po \cite{Besl1991}                   & 0.139 & 1.176  & 0.089  & 2.017   \\
			ICP-Po2Pl \cite{Besl1991}                   & 0.084 & 1.693  & 0.065  & 2.050   \\
			G-ICP \cite{Segal2009}               & \textbf{0.067} & \textbf{0.375}  & \textbf{0.065}  & 2.045   \\
			AA-ICP \cite{Pavlov2018}             & 0.145 & 1.406  & 0.088  & 2.020   \\
			NDT-P2D \cite{Todor2012}             & 0.101 & 4.369  & 0.071  & 2.000   \\
			CPD \cite{Myronenko2010}             & 0.461 & 5.076  & 0.804  & 7.301   \\
			3DFeat-Net \cite{yew20183dfeat}            & 0.199 & 2.428  & 0.116  & 4.972   \\
			\midrule[.5pt]
			Ours-Base                                & 0.195  & 1.700  & 0.073   & 0.482   \\
			Ours-Duplication                                & 0.164  & 1.212  & 0.071  & \textbf{0.482}   \\
			\bottomrule[1pt] 
		\end{tabular}
	\end{center}
	\vspace{-0.2cm}
	\caption{
		Comparison using the KITTI dataset. Our performance is comparable against traditional geometry-based methods and better than the learning-based method, 3DFeat-Net. The much lower maximum errors demonstrate good robustness.
	}
	\label{tab:kitti_data_result}
	\vspace{-0.4cm}
\end{table}

\vspace{0.2cm}
\textbf{Apollo-SouthBay Dataset}
In Apollo-SouthBay dataset, we sample at $100$ frame intervals, and again enumerate the target within $5m$ distance. All other parameter settings for each individual method are the same as the KITTI dataset.
The angular and translational errors are listed in Table~\ref{tab:our_data_result}.
For the Apollo-SouthBay dataset, most methods including ours have a performance improvement, which might be due to the better ground truth poses provided by the dataset.
Our system with the duplication design achieves the second-best mean translational accuracy and comparable angular accuracy with regards to other traditional methods. Additionally, the lowest maximum translational error demonstrates good robustness and stability of our proposed learning-based method.

\begin{table}[htbp]
	\small
	\begin{center}
		\begin{tabular}{p{2.3cm}|C{1.0cm}C{1.0cm}|C{1.0cm}C{1.0cm}}
			\toprule[1pt]
			\multirow{2}{*}{\makecell[tl]{Method}} &\multicolumn{2}{c}{Angular Error($^\circ$)} &\multicolumn{2}{c}{Translation Error($m$)}  \\
			\cline{2-3}\cline{4-5}
			& Mean & Max & Mean & Max  \\
			\midrule[.5pt]
			ICP-Po2Po \cite{Besl1991}                   & 0.051 & 0.678  & 0.089  & 3.298   \\
			ICP-Po2Pl \cite{Besl1991}                   & 0.026 & \textbf{0.543}  & 0.024  & 4.448   \\
			G-ICP \cite{Segal2009}               & \textbf{0.025} & 0.562  & \textbf{0.014}  & 1.540   \\
			AA-ICP \cite{Pavlov2018}             & 0.054 & 1.087  & 0.109  & 5.243   \\
			NDT-P2D \cite{Todor2012}             & 0.045 & 1.762  & 0.045  & 1.778   \\
			CPD \cite{Myronenko2010}             & 0.054 & 1.177  & 0.210  & 5.578   \\
			3DFeat-Net \cite{yew20183dfeat}             & 0.076 & 1.180  & 0.061  & 6.492   \\
			\midrule[.5pt]
			Ours-Base                                & 0.135  & 1.882  & 0.024   & \textbf{0.875}   \\
			Ours-Duplication                                & 0.056  & 0.875  & 0.018   & 0.932  \\
			\bottomrule[1pt] 
		\end{tabular}
	\end{center}
	\vspace{-0.2cm}
	\caption{
		Comparison using the Apollo-SouthBay dataset. Our system achieves the second best mean translational error and the lowest maximum translational error. The low maximum errors demonstrate good robustness of our method.
	}
	\label{tab:our_data_result}
	\vspace{-0.2cm}
\end{table}

%
%

\vspace{0.2cm}
\textbf{Run-time Analysis}
We evaluate the runtime performance of our framework with a GTX 1080 Ti GPU, Core i7-9700K CPU, and 16GB Memory as shown in Figure~\ref{fig:runtime}.
The total end-to-end inference time of our network is about $2$ seconds for registering a frame pair with the duplication design in Section~\ref{section:impl}.
Note that DeepICP is significantly faster than the other learning-based approach, 3DFeat-Net \cite{yew20183dfeat}, because we extract only $64$ keypoints instead of $1024$, and do not rely on a RANSAC procedure.

\vspace{-0.2cm}

\subsection{Ablations}
\label{subsec:ablation}

In this section, we use the same training and testing data from the Apollo-SouthBay dataset to further evaluate each component or proposed design in our work.

\vspace{0.2cm}
\textbf{Deep Feature Embedding}
In Section~\ref{subsec:dfe}, we propose to construct the network input by concatenating the FE feature together with the local coordinates and the intensities of the neighboring points. Now, we take a deeper look at this design choice by conducting the following experiments: i) LLF-DFE: Only the local coordinates and the intensities are used; ii) FEF-DFE: Only the FE feature is used; iii) FEF: The DFE layer is discarded. The FE feature is directly used as the input to the CPG layer. In the target point cloud, the FE features of the grid voxel centers are interpolated.
It is seen that the DFE layer is crucial to this task as there is severe performance degradation without it as shown in Table~\ref{tab:dfe_ablation}.
The LLF-DFE and FEF-DFE give competitive results while our design gives the best performance.

\begin{table}[htbp]
	\small
	\begin{center}
		\begin{tabular}{p{1.7cm}|C{1.1cm}C{1.1cm}|C{1.1cm}C{1.1cm}}
			\toprule[1pt]
			\multirow{2}{*}{\makecell[tl]{Method}} &\multicolumn{2}{c}{Angular Error($^\circ$)} &\multicolumn{2}{c}{Translation Error($m$)}  \\
			\cline{2-3}\cline{4-5}
			& Mean & Max & Mean & Max  \\
			\midrule[.5pt]
			LLF-DFE               & 0.058 & 0.861  & 0.024  & 0.813   \\
			FEF-DFE               & 0.057 &  \textbf{0.790}  & 0.026  & \textbf{0.759}   \\
			FEF                   & 0.700 & 2.132  & 0.954  & 8.416   \\
			Ours                 &  \textbf{0.056}  & 0.875  &  \textbf{0.018}   &  0.932   \\
			\bottomrule[1pt] 
		\end{tabular}
	\end{center}
	\vspace{-0.2cm}
	\caption{
		Comparison w/o the DFE layer. The usage of DFE layer is crucial as there is severe performance degradation as shown in Method FEF. When only partial features are used in DFE layer, it gives competitive results as shown in Method LLF-DFE and FEF-DFE, while ours yields the best performance.
	}
	\label{tab:dfe_ablation}
	\vspace{-0.2cm}
\end{table}

\vspace{0.2cm}
\textbf{Corresponding Points Generation}
To demonstrate the effectiveness of the CPG, we directly search the best corresponding point among the existing points in the target point cloud taking the predicted transformation into consideration.
Specifically, for each source keypoint, the point with the highest similarity score in the feature space in the target neighboring field is chosen as the corresponding point.
It turns out that it is unable to converge using our proposed loss function.
The reason might be that the proportion of the positive and negative samples is extremely unbalanced.

\begin{figure}[htbp]
	\centering
	\includegraphics[width=\linewidth]{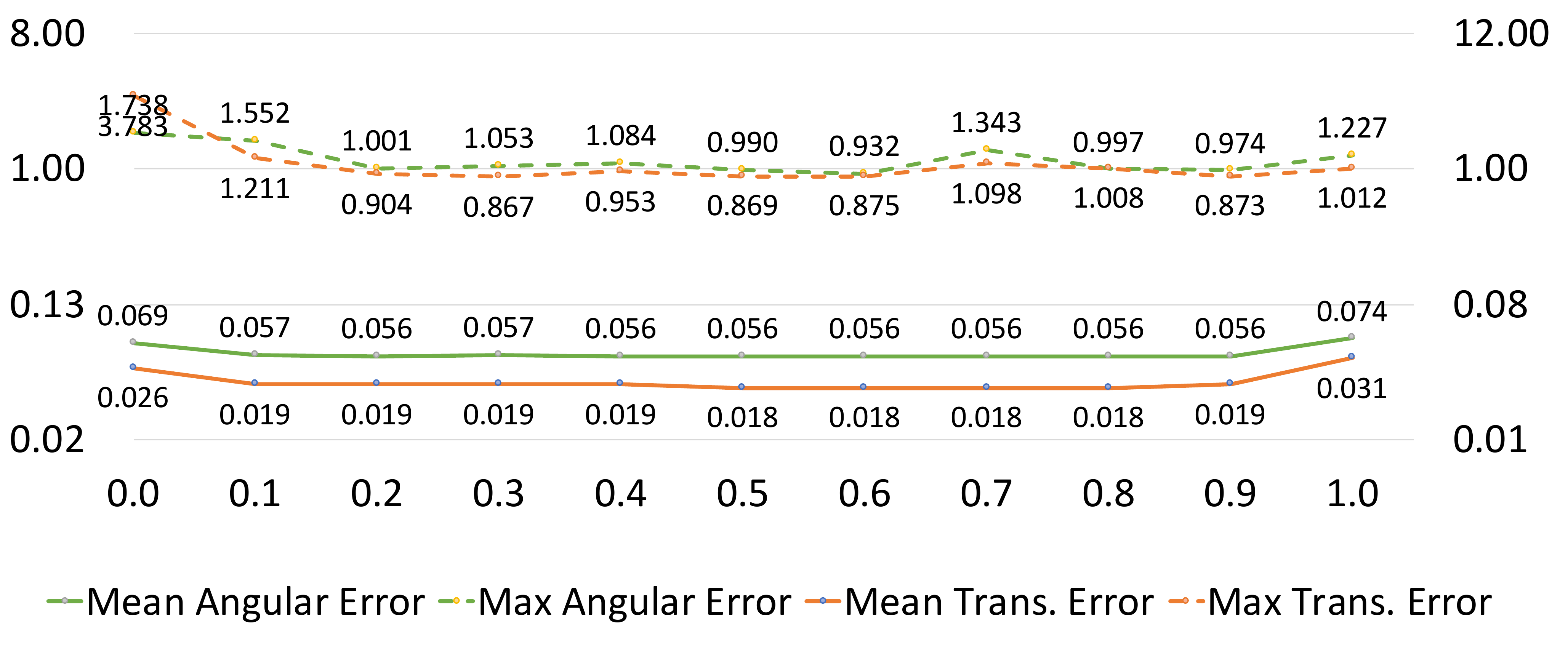}
	\vspace{-0.5cm}
	\caption{
		Registration accuracy comparison with different $\alpha$ values in the loss function. Any $\alpha$ values away from $0.0$ or $1.0$ give similarly good accuracies. This demonstrates the powerful generalization capability of deep neural networks.
	}
	\label{fig:loss_ablation}
	\vspace{-0.4cm}
\end{figure}

\vspace{0.2cm}
\textbf{Loss}
In Section~\ref{subsec:loss}, we propose to use the combination of two losses to incoorporate the global geometric information, and a balancing factor $\alpha$ is introduced.
In order to demonstrate the necessity of using both the losses, we sample $11$ values of $\alpha$ from $0.0$ to $1.0$ and observe the registration accuracy.
In Figure~\ref{fig:loss_ablation}, we find that the balancing factor of $0.0$ and $1.0$ obviously give larger angular and translational mean errors. This clearly demonstrates the effectiveness of the combined loss function design. It is also quite interesting that it yields similar accuracies for $\alpha$ between $0.1$ - $0.9$. We conclude that this might be because of the powerful generalization capability of deep neural networks. The parameters in the networks can be well generalized to adopt any $\alpha$ values away from $0.0$ or $1.0$. Therefore, we use $0.6$ in all our experiments.

\begin{figure*}[htbp]
	\centering
	\includegraphics[width=\linewidth]{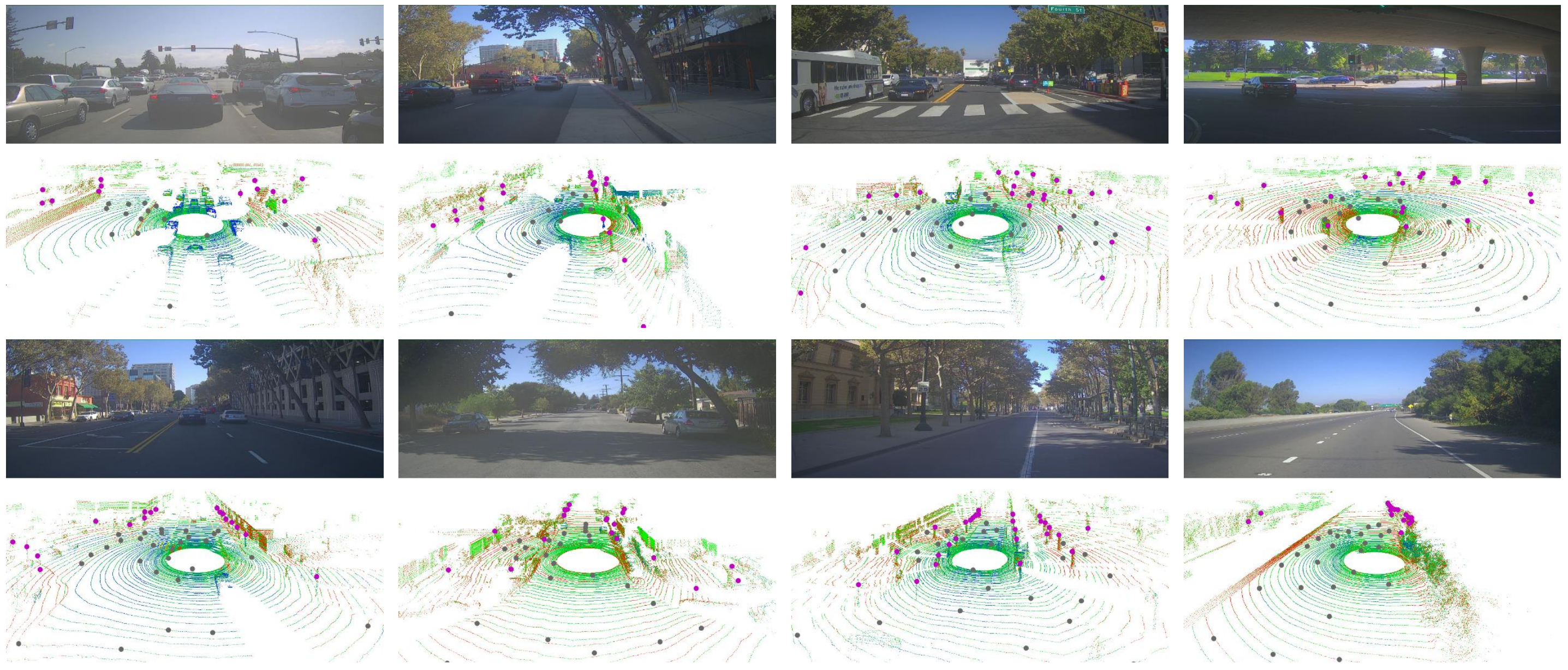}
	\vspace{-0.3cm}
	\caption{
		Visualization of the detected keypoints by the point weighting layer. The pink and grey keypoints are detected by the front and back network, respectively. The pink ones appear on stationary objects, such as tree trunks and poles. The grey ones are mostly on the ground, as expected.
	}
	\label{fig:weightingpoints}
	\vspace{-0.1cm}
\end{figure*}

\begin{figure*}[htbp]
	\centering
	\includegraphics[width=\linewidth]{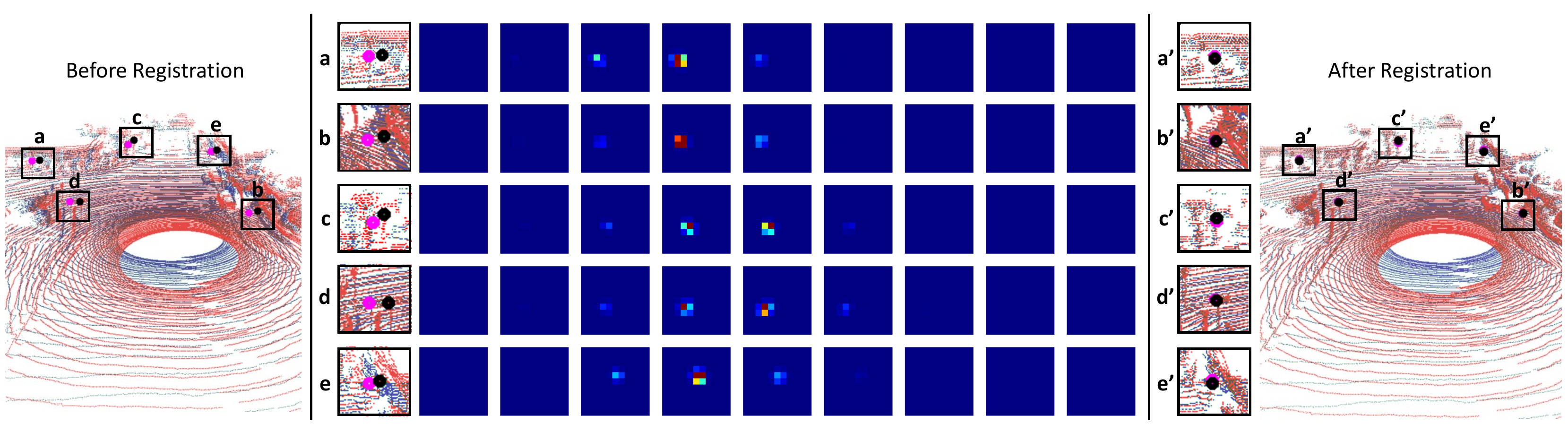}
	\vspace{-0.3cm}
	\caption{
		Illustrate the matching similarity probabilities of each keypoint to its matching candidates by visualizing them in $x$ and $y$ dimensions with $9$ fixed $z$ values. The black and pink points are the detected keypoints in the source point cloud and the generated ones in the target, respectively. The effectiveness of the registration process is shown on the left (before) and right (after).
	}
	\label{fig:cpg_distribution}
	\vspace{-0.2cm}
\end{figure*}

\subsection{Visualizations}
\label{subsec:visulaizations}

In this section, to offer better insights on the behavior of the network, we visualize the keypoints chosen by the point weighting layer and the similarity probability distribution estimated in the CPG layer.

\vspace{0.2cm}
\textbf{Visualization of Keypoints}
In Section~\ref{subsec:fl}, we propose to extract semantic features using PointNet++ \cite{qi_2017_pointnetplusplus}, and weigh them using a MLP network structure.
We expect that our end-to-end framework can intelligently learn to select keypoints that are unique and stable on stationary objects, such as traffic poles, tree trunks, but avoid the keypoints on dynamic objects, such as pedestrians, cars.
In addition to this, we duplicate our network in Section~\ref{section:impl}. The front network with the 3D CNNs CPG layer are expected to find meaningful keypoints those have good constraints in all six degrees of freedom. While the back network with the 1D CNNs are expected to find those are good in $z$, $roll$ and $pitch$ directions.
In Figure~\ref{fig:weightingpoints}, the detected keypoints are shown compared with the camera photo and the LiDAR scan in the real scene.
The pink and grey keypoints are detected by the front and back network, respectively.
We observe that the distribution of keypoints match our expectations as the pink keypoints mostly appear on objects with salient features, such as tree trunks and poles, while the grey ones are mostly on the ground.
Even in the scene where there are lots of cars or buses, none of keypoints are detected on them.
This demonstrates that our end-to-end framework is capable to detect the keypoints those are good for the point cloud registration task.

\vspace{0.2cm}
\textbf{Visualization of CPG Distribution}
The CPG layer in Section~\ref{subsec:corrs} estimates the matching similarity probability of each keypoint to its candidate corresponding ones.
Figure~\ref{fig:cpg_distribution} depicts the estimated probabilities by visualizing them in $x$ and $y$ dimensions with $9$ fixed $z$ values.
On the left and right, the black and pink points are the keypoints from the source point cloud and the generated ones in the target, respectively.
It is seen that the keypoints detected are sufficiently salient that the matching probabilities are concentratedly distributed.

\section{Conclusion}
\label{section:conclusion}

We have presented an end-to-end framework for the point cloud registration task. 
The keypoints are detected through a point weighting deep neural network.
The corresponding point to the keypoint is generated according to the matching similarity probabilities estimated by a 3D CNNs structure among candidates, but isn't directly picked from the existing ones.
Our loss function incoporates both the local similarity and the global geometric constraints.
These novel designs in our network make our learning-based system achieve the comparable registration accuracy to the state-of-the-art geometric methods.
The LiDAR point cloud registration is an important task, and is the foundation of various applications.
We believe this has great benefits for many potential applications.
In a further extension of this work, we will develop the potential of our system with more LiDAR models and application scenarios.

{\small
\bibliographystyle{ieee}
\bibliography{ms}
}

\end{document}